\pdfoutput=1
\documentclass[10pt, a4paper]{article}
\usepackage[]{lrec-coling2024} % this is the new style

\usepackage{microtype}

\usepackage{type1cm}
\usepackage{url}
\usepackage{amssymb} 
\usepackage{amsmath}
\usepackage{multirow} 
\usepackage{xspace}
\DeclareMathAlphabet{\mathcalbf}{OMS}{pzc}{b}{n}
\usepackage{microtype}

\usepackage{tabularx}
\usepackage{booktabs}
\setkeys{Gin}{pagebox=artbox}

\newcommand{\hwfigure}[3][t!]{%
	\begin{figure*}[#1]%
		\centering%
		\includegraphics[scale=1.0]{#2}%
    		\caption{#3}\label{#2}%
  	\end{figure*}
}

\RequirePackage{type1cm}
\RequirePackage{color}
\RequirePackage{soul}
\setstcolor{blue}
\definecolor{violet}{rgb}{0.5,0.0,0.5}
\newsavebox\bscombox
\newcommand{\bscom}[3][]{%
	\sbox{\bscombox}{\fontsize{8}{9}\selectfont#1#2#3}
	\noindent
	\st{#2}{\selectfont
		\color{blue}#3\ifx\\#1\\\else{\fontsize{8}{9}\selectfont\color{violet}[#1]}\fi
	}
}

\definecolor{highlight1}{rgb}{0.95,0.95,0.95}
\definecolor{tgray}{rgb}{0.5,0.5,0.5}
\definecolor{tgray}{rgb}{0.5,0.5,0.5}

\title{Argument Quality Assessment \\ in the Age of Instruction-Following Large Language Models}

\name{Henning Wachsmuth,$^{1}$ Gabriella Lapesa,$^{2}$ Elena Cabrio,$^{3}$ Anne Lauscher,$^{4}$ \\ 
	{\bf \large Joonsuk Park,$^{5}$ Eva Maria Vecchi,$^{6}$ Serena Villata,$^{3}$ Timon Ziegenbein$^{1}$}}

\address{%
$^{1}$Leibniz University Hannover, \{h.wachsmuth,t.ziegenbein\}@ai.uni-hannover.de \\
$^{2}$GESIS, Heinrich-Heine University D\"usseldorf, gabriella.lapesa@gesis.org \\
$^{3}$Universit\'e C\^ote d'Azur, CNRS, Inria, I3S \{elena.cabrio,serena.villata\}@univ-cotedazur.fr \\
$^{4}$Universit\"at Hamburg, anne.lauscher@uni-hamburg.de \\
$^{5}$University of Richmond, park@joonsuk.org \\
$^{6}$University of Stuttgart, eva-maria.vecchi@ims.uni-stuttgart.de\\
}

\abstract{
The computational treatment of arguments on controversial issues has been subject to extensive NLP research, due to its envisioned impact on opinion formation, decision making, writing education, and the like. A critical task in any such application is the assessment of an argument's quality---but it is also particularly challenging. In this position paper, we start from a brief survey of argument quality research, where we identify the diversity of quality notions and the subjectiveness of their perception as the main hurdles towards substantial progress on argument quality assessment. 
We argue that the capabilities of instruction-following large language models (LLMs) to leverage knowledge across contexts enable a much more reliable assessment. Rather than just fine-tuning LLMs towards leaderboard chasing on assessment tasks, they need to be instructed systematically with argumentation theories and scenarios as well as with ways to solve argument-related problems. We discuss the real-world opportunities and ethical issues emerging thereby.
\\ \newline \Keywords{Computational Argumentation, Argument Quality, Large Language Model, Instruction Fine-Tuning} 
}

\begin{document}
\maketitleabstract
\section{Introduction}
\label{sec:introduction}

\emph{``In some sense, the question about the quality of an argument is the `ultimate' one for argumentation mining.''} \cite{stede:2018}.

\medskip
When learning about controversial issues, people rarely accept arguments they encounter without further contemplation. Rather, they seek to find \emph{the best} arguments; those that help them form an opinion or write texts that persuade others; those that make them reach agreement or at least understand each other better. That is to say, \emph{argument quality} is of interest as soon as arguments are presented to an audience. Computational argumentation aids the treatment of arguments at a larger scale, with important applications in search \cite{wachsmuth:2017e}, business \cite{slonim:2021}, and education \cite{wambsganss:2022}. But the situation there is the same: It is not enough to mine or generate arguments; their quality also needs to be evaluable \cite{park:2018}, so that it can be assessed \cite{lauscher:2020}, flaws can be found \cite{goffredo:2022}, and accounted for \cite{skitalinskaya:2023b}.

\citet{wachsmuth:2017b} surveyed research on argument quality assessment, organizing theories and methods under 15 quality notions, from logical cogency to rhetorical effectiveness to dialectical reasonableness. Even though computational argumentation was just gaining momentum in natural language processing (NLP) back then, rarely going beyond argument mining, two inherent challenges of argument quality were visible already: the \emph{diversity} of quality notions as well as the \emph{subjectivity} of their perception and, hence, of their assessment for both humans and computational models. Consider the following argumentative claim against censoring Mark Twain's usage of the N-word, taken from the debate platform kialo.com:

\medskip \noindent
\emph{``In Huckleberry Finn, Twain captured the essence of everyday midwest American English."}

\medskip 
This claim is certainly relevant to the discussion, but whether people will deem it effective may strongly depend on their individual context. A person without African-American background may be willing to accept the argument; one with high literacy might look for clearer logical connections.

While the challenges of diversity and subjectivity prevail until today \cite{lapesa:2023}, NLP is now seeing a revolutionary breakthrough: the rise of \emph{instruction-following large language models} (henceforth, LLMs) that can tackle various NLP tasks with little to no task-specific fine-tuning, enabled by their supreme capability to integrate and leverage knowledge across contexts \cite{openai:2023}. The question is: \emph{What are the implications for argument quality assessment specifically as well as for computational argumentation in general?}

In this position paper, we revisit the computational assessment of argument quality in light of the availability of LLMs such as GPT-4 and Alpaca \cite{taori:2023}. Starting from the status quo reported by \citet{wachsmuth:2017b}, we carry out a brief survey of recent NLP research on the topic (Section~\ref{sec:relatedwork}). To bring order into the various lines of research pursued since 2017, we organize them into three general directions, as laid out in~Figure~\ref{quality-research.pdf}: 
\begin{itemize}
\setlength{\itemsep}{0pt}
\item
\emph{Conceptual notions} of maximal and minimal argument quality, 
\item
\emph{Influence factors} of argument quality from the context where arguments occur, and
\item
\emph{Computational models} for assessing or improving argument quality.
\end{itemize}

On this basis, we establish the central question to which we provide answers in this paper:
\begin{quote}
\emph{How to drive research on LLM-based argument quality assessment in order to face the prevailing challenges of diverse quality notions and their subjectivity?}
\end{quote}

In particular, we are convinced that the capabilities of instruction-following LLMs enable research to overcome many aspects of the two challenges. To this end, the primary focus of NLP research on argument quality should be put on systematic ways to teach LLMs to follow instructions, including concepts and settings of arguing in addition to ways to solve argument-related problems (Section~\ref{sec:model}). Instead of fine-tuning LLMs on predefined domains (manifested in the training data) and preselected theories (manifested in the data's annotations), as well as simple engineering of prompts, we expect the greatest impact to lie in teaching LLMs the theories, circumstances, and ethical constraints to adhere to. The rationale behind this is that LLMs will often have processed data from all contexts needed to make an informed judgment about an argument's quality, due to their heavy pretraining on huge amounts of data. In contrast, LLMs cannot access, by default, the knowledge of what is to be prioritized in a given setting. 
 
We state upfront that the blueprint delineated in this paper comes with several limitations and ethical considerations that we critically analyze below. Moreover, we are naturally aware of the general issues of LLMs, including hallucinated facts and the reproduction of common social biases. These issues deserve treatment in computational argumentation as well; they are even particularly critical there due to the sensitivity of many controversial topics \cite{holtermann:2022}. Keeping this in mind, we believe that it is necessary to explore now how to best employ LLMs for argument quality assessment in order to harness their full potential for the main applications, while avoiding to waste energy for the typical pursuit of leaderboard rankings on existing quality assessment tasks. This is the goal of the paper at hand.

Now, why is it important to discuss LLMs for argument quality assessment specifically? We address this matter when we look at the real-world opportunities emerging from the capabilities of LLMs in academia and industry (Section~\ref{sec:analysis}). While \newcite{argyle:2023} developed LLMs that tone down argumentative conversations, we postulate a contrary path: Exploiting the means of LLMs to proactively enable people to learn and better reason about controversial issues, thus contributing towards more deliberate conversations \cite{vecchi:2021}. We think that the time has come to revisit and pursue the core visions of computational argumentation research, from the overcoming of filter bubbles to the individualized mass education of learners. We sketch how these visions could be realized with the LLMs available today, before we conclude (Section~\ref{sec:conclusion}) and stress ethical concerns that arise with LLMs that actively affect human views (Section~\ref{sec:limitations}).\,\,

With the discussion in this position paper, we provide two main contributions to research:
\begin{enumerate}
\setlength{\itemsep}{0pt}
\item
\emph{A survey} of the main lines of recent research on argument quality and its assessment
\item
\emph{A blueprint} for impactful future research on LLMs for argument quality assessment 
\end{enumerate}
 % About 1.50 pages including title+abstract
\section{A Brief Survey of Recent Research}
\label{sec:relatedwork}

To start, this section briefly but systematically surveys recent NLP work on argument quality assessment. We identify three general directions, each with two main aspects, and organize the research accordingly, as illustrated in Figure~\ref{quality-research.pdf}.

\hwfigure{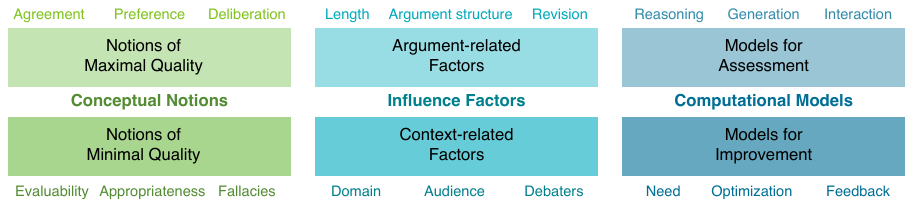}{Organization of the surveyed argument quality research into three general directions (conceptual notions, influence factors, and computational models), their main aspects (e.g., notions of maximal and minimal quality), and specific concepts studied for these (e.g., agreement, preference, and deliberation).}

\subsection{Frame of the Survey}

Beyond holistic computational argumentation (CA) surveys \cite{stede:2018,cabrio:2018,lawrence:2019}, \newcite{wang:2023b} specifically reviewed works on argument generation, \newcite{lauscher-etal-2022-scientia} on knowledge in CA,  and \newcite{vecchi:2021} on the use of CA for the social good. Also, some tutorials treated argument mining and its applications \cite{%
% NOTE (HW). Too early
%%%slonim:2016,
budzynska:2019,barhaim:2021}. In contrast, we focus on argument quality assessment.

Aside from our recent tutorial \cite{lapesa:2023}, the only argument quality review that we are aware of is the one of \citet{wachsmuth:2017b} who organize relevant literature until mid 2017 into a taxonomy of 15 logical, rhetorical, and dialectical quality dimensions. The authors already discussed the diversity of quality notions as well as the subjectivity of their perception, both of which are still hampering research. In this paper, we seek to delineate ways to overcome them, in line with the authors' organization of what argument quality means. We start from their work here, so we restrict our survey to work that is published after theirs. 

Based on our experience with NLP research on CA, we cover four groups of publication venues:
\begin{itemize}
\setlength{\itemsep}{0pt}
\item
All NLP venues covered by the ACL anthology
\item
The leading artificial intelligence (AI) conferences, all from AAAI.org and IJCAI
\item
The leading information retrieval (IR) conferences, SIGIR and ECIR
\item
The leading CA conference, COMMA
\end{itemize}

We used Google site search and Springer and ACM's internal search on September 1, 2023 (updated on October 17, 2023), to gather all papers containing any of the following pairs of words:
\begin{eqnarray*}
&& \{\textit{argument, argumentation, debate}\} \\
& \times & \{\textit{quality, strength, persuasiveness}\}
\end{eqnarray*}

This led to 257 papers (202~NLP, 35~AI, 11~IR, 9~CA). From these, we kept all 119 papers that deal with quality of natural language arguments based on title, abstract, and skimming (98 NLP, 12 AI, 6 IR, 3 CA). To focus on scientific novelty, we further excluded surveys, tutorials, demos, shared tasks, and system papers, leaving 104 papers (87 NLP, 10 AI, 5 IR, 2 CA). We checked these in more detail to ensure that argument quality is actually part of the research. After filtering out others, we obtained a final set of 83 papers (69 NLP, 9 AI, 3 IR, 2 CA).\,\,

\subsection{General Research Directions}

Analyzing the 83 papers as a whole, we identified the following three general directions of research on argument quality, each with two main aspects. Concretely, one author of this paper proposed the organization. The papers were then distributed among all other authors who ranked them by the directions and aspects they contribute, if any.

\paragraph{Conceptual Notions} 
24 of the papers primarily deal with the question of what is actually meant by argument quality, considered from either of two complementary perspectives:
\begin{itemize}
\setlength{\itemsep}{0pt}
\item
\emph{Notions of maximal quality} based on arguing goals such as agreement and deliberation or on preferences between different arguments
\item
\emph{Notions of minimal quality} in terms of what makes an argument evaluable or appropriate to be stated as well as how to avoid fallacies
\end{itemize}

\paragraph{Influence Factors} 
This direction covers 30 papers studying (or controlling) two types of factors that influence the perception of quality beyond the content, structure, and style of the argument itself:
\begin{itemize}
\setlength{\itemsep}{0pt}
\item
\emph{Argument-related factors} such as the argument's length, its structure in terms of relations between units, and revisions applied to it
\item
\emph{Context-related factors}, such as the domain of the discussion, the audience addressed, and the debaters involved
\end{itemize}

\paragraph{Computational Models} 
Finally, 21 papers aim mainly at methodological novelty in the modeling of argument quality for two quality-related tasks:
\begin{itemize}
\setlength{\itemsep}{0pt}
\item
\emph{Models for assessment} of argument quality, capturing specificities of the task, the whole discussion, or the context of arguing
\item
\emph{Models for improvement} of argument quality, targeting the need for improvement, actual optimizations, or feedback on what to improve
\end{itemize}

\medskip 
The remaining eight papers pursue individual research directions. 
We note that many of the surveyed papers do not fall under one general direction only; rather, they often have a visible focus on one of them. In particular, our internal discussion revealed that contributions to influence factors and computational models are not always easy to distinguish and that, sometimes, models may rather target downstream applications. Still, the directions and aspects were agreed upon in general.%
\footnote{For validation, we reassigned 16 papers (19\%) to other authors: 11 got the same main direction; for four, it was seen as the second contribution. Only in one case, a fully different direction was assigned. After rechecking, the newly assigned direction did not seem adequate.}

In the following, we discuss selected works from each of the general research directions. Table~\ref{table-survey} in the appendix shows the full list of all 83 covered publications, grouped by the primary general research direction and the main aspect.

\subsection{Conceptual Notions}

Naturally, all surveyed literature builds on some notion of argument quality, at least implicitly. However, we found that 24 of the 83 papers have the explicit treatment of quality notions as their main focus and 10 further papers contribute to quality notions to some extent. About two-thirds of the works discuss how an argument should be ideally (maximal quality), the others what an argument should at least achieve or avoid (minimal quality).

\paragraph{Notions of Maximal Quality}

Some researchers build on the argument quality taxonomy proposed by \newcite{wachsmuth:2017b}, including \newcite{lauscher:2020} who model the main taxonomy notions using multitask learning across Q\&A, debate, and review forums. Others question the simplifying view that argument quality is about persuasion only: \newcite{elbaff:2018} consider the goal of \emph{agreement}, defining good news arguments as those that challenge or corroborate stance. \newcite{gretz:2020} see argument quality as a \emph{preference} relation, and \newcite{falk:2021} examine its connection to \emph{deliberation}. With an entirely different perspective, some papers examine what makes an argument good irrespective of topic \cite{beigman-klebanov:2017}, whereas, for example, \newcite{dumani:2020} operationalize argument quality for practice in a quality-based framework for argument retrieval.

\paragraph{Notions of Minimal Quality}

\newcite{park:2015} establish the notion of an argument's \emph{evaluability}, that is, the prerequisite of assessing logical quality soundly. A key research line on minimal quality is the detection of \emph{fallacies}: arguments with flawed or deceptive reasoning. Neural models have shown success on this deep semantic problem; some aim at ad-hominem arguments only \cite{habernal:2018}, others at various fallacies \cite{jin:2022}. \newcite{persing:2017} tackle the broader problem of spotting an argument's weaknesses, from grammar errors to lack of objectivity and unclear justifications. More practice-oriented, \newcite{pauli:2022} look at the misuse of fallacies for rhetorical appeals in online forums and fake news. Finally, \newcite{ziegenbein:2023} refine the notion of \emph{appropriateness}, emanating from Aristotle's work \cite{aristotle:2007}. They see it as the minimal quality that makes arguments worthy of being considered and annotate data for violations of appropriateness.

\subsection{Influence Factors}

Assessing the different notions of argument quality is a complex task and is influenced by many factors, some of which have no explicit relation to the argument itself. Accordingly, research has dealt with the identification, modeling, and controlling of such factors and their impact on argument quality. We found that 30 of the 83 papers mainly focus on influence factors, and a further 19 papers are to some extent devoted to them. Of these, about 60\% discuss argument-related factors while the rest looks at context-related factors.

\paragraph{Argument-related Factors}

In terms of textual factors influencing the perceived quality of arguments, researchers display the questionable power of \emph{length} as a predictor \cite{potash:2017} and account for this in dataset creation \cite{toledo:2019}. The impact on quality of internal \emph{argument structure} has been investigated using the notion of organization quality in learner essays \cite{chen:2022} and by using annotations of argument components in business model pitches \cite{wambsganss:2022}. Notions of structure within an argument are further extended through adding attributes to argument components \cite{carlile:2018}, shifting the focus to component-related factors, or by comparing different \emph{revisions} of the same claim \cite{skitalinskaya:2021}.

\paragraph{Context-related Factors}
 
\newcite{lukin:2017} analyze the interaction between argumentative styles (emotional vs.\ factual) and the personality of the \emph{audience}, as modeled by the Big Five traits. Similarly, \newcite{durmus:2018} model political and religious ideologies, based on the audience's stances on various controversial topics. Both indicate that audience-level factors often outweigh language use in their persuasive effect. \newcite{alshomary:2022} turn the view to rhetorical strategies of \emph{debaters}, assessing the effect of morally-framed arguments. They find that morals are particularly successful in challenging the audience's beliefs. \newcite{wiegmann:2022} analyze stylistic and behavioral characteristics of debaters that contribute to their persuasiveness over multiple debates. Aside from debate participants, \newcite{liu:2022} explore arguments on social media that are accompanied by images, highlighting the potential of multimodal approaches to quality assessment, whereas \newcite{fromm:2023} generalize the contextual scope of assessment to multiple \emph{domains} at the same time.\,\,\,

\subsection{Computational Models}

The majority of the 83 surveyed papers include empirical experiments with models for argument quality. However, we found that only 21 of them actually focus on proposing novel approaches targeting either of the above-mentioned conceptual quality notions, whereas 26 other papers have such approaches as secondary contributions to support their claims with experimental results and analysis. Almost all approaches aim at the assessment of argument quality, but a few recent ones go beyond assessment, studying how to improve quality.

\paragraph{Models for Assessment}

Many approaches aim at specific quality notions. For example, the attentive interaction model of  \newcite{jo:2018} predicts an opinion holder's view change by detecting vulnerable regions in their \emph{reasoning} and modeling its relation to a challenger's argument. \newcite{gleize:2019} propose a Siamese neural network to assess the convincingness of evidence, while \newcite{song:2020} develop a hierarchical multitask learning approach to jointly model discourse element identification and organization assessment for essay scoring. \newcite{gurcke:2021} examine to what extent an argument's logical sufficiency can be predicted based on whether its conclusion can be inferred from its premises using the \emph{generation} capabilities of transformers. \newcite{kondo:2021} assess the validity of an argument's reasoning using Bayesian networks and predicate logic facilitated by argumentation schemes. A few works also look beyond single quality notions, such as \newcite{falk:2023} who inject knowledge about the \emph{interactions} between different quality notions to improve the prediction of individual ones. 

\hwfigure{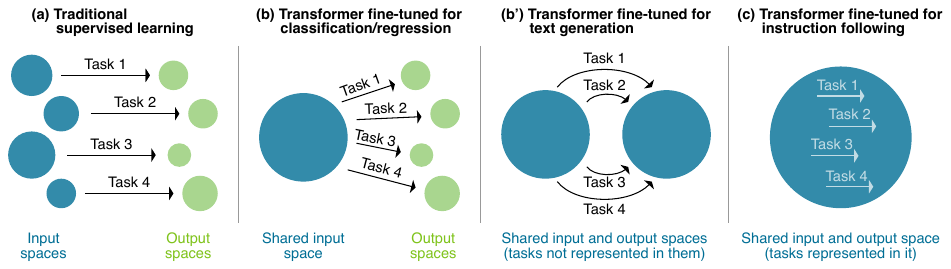}{Learning of representational spaces in NLP models (same color: same type of representation): \emph{(a)~Traditional supervised learning:} Input and output spaces are separated across tasks; representations are task-specific. \emph{(b)~Classification/Regression transformer:} The input space is shared~across~tasks;~its representation can be learned on all tasks. \emph{(b')~Generation transformer:} Both spaces are shared across tasks; their representations can be learned on all tasks, but not task interactions. \emph{(c)~Instruction-following transformer:} One space for inputs, outputs, and tasks; representations can be learned jointly on all tasks.}%

\paragraph{Models for Improvement}

While only a few models for improvement have been presented so far, we expect more to come soon, also seeing related efforts on topics beyond those covered in this survey \cite{chakrabarty:2021,ihori:2022,li:2022}. An early attempt was made by \newcite{ke:2018} who design neural models that predict the persuasiveness and other attributes of arguments in a student essay, to provide \emph{feedback} to students on how to improve their arguments. Recently, \newcite{skitalinskaya:2023} identified arguments in \emph{need} of improvement, leveraging complex revision-based data with transformer models. \newcite{skitalinskaya:2023b} go one step further, presenting the first approach to the \emph{optimization} of argumentative claims. It combines neural claim rewriting with quality-based ranking.

\subsection{Other Research Directions}

Among the eight papers that do not fit under the three main research directions, we identified the following two rough research areas.

Five papers deal with specific applications for which argument quality assessment is key. \newcite{rach:2020} and \newcite{kiesel:2020} target argument search, both taking a human-interaction perspective: The former studies the effects of integratting argument search into an avatar-based dialogue system; the latter investigates user expectations on voice-based argument search systems, such as preferred ranking criteria. \citet{chalaguine:2020} develop a chatbot that relies on an argument graph for persuasive counterargument generation, and \citet{falk:2021} address expert moderation in a deliberative forum, taking moderator interventions as implicit labels for the need to improve comment quality. \citet{fromm:2021} use argument mining for analyzing peer reviews.

The other works tackle the bottleneck of (scarce) assessment training data. \citet{heinisch:2022} employ data augmentation to support the prediction of argument validity and novelty. \citet{kees:2021} evaluate active learning strategies for supporting argument strength estimation, and \citet{yang:2019} introduce a quality control method that they apply to annotate argument acceptability.

 % About 3.00 pages
\section{LLMs for Argument Quality}
\label{sec:model}

Section~\ref{sec:relatedwork} stresses that a big part of argument quality research tackles the key challenges of diverse views of quality (by developing or refining notions) and subjectivity (by controlling or modeling influence factors). However, the intricated interdependencies between different quality notions and the various factors that influence them have hampered substantial progress in the reliable assessment of argument quality so far. We argue that instruction-following large language models (LLMs) have the potential to overcome many limitations, if systematic ways to teach them accordingly are established. In this section, we start from the main advantages of such LLMs. Then, we outline what to instruct LLMs with and how to do so (beyond simple in-context learning/prompting techniques) in order to advance LLMs for argument quality in future research.

\subsection{Assessment without Instructions} 

Conceptually, argument quality assessment is a classification or regression problem, even if partly treated as preference learning. For decades, research in NLP relied on the traditional supervised learning paradigm for most such problems: to induce a mapping from one representational space to another using training pairs of input and output. As sketched in Figure~\ref{model-concepts.pdf}a, the input spaces (usually, representing natural language) and output spaces (label schemes or value ranges, such as argument quality scores) are separated thereby. This separation prevents any exchange of knowledge across spaces and across tasks.

With the shift to transformers in NLP, the learning effort is mostly reduced to the self-supervised pretraining of a language model \cite{vaswani:2017,devlin:2019}. Under the transfer learning paradigm, only fine-tuning remains supervised, to make a model address the task it is supposed to. This way, knowledge is shared between input representations across tasks and contexts, that is, all texts ever processed affect how an input is encoded. In classification and regression, however, fine-tuning (say, of a BERT encoder~with a quality scoring head) reintroduces a key restriction of traditional methods, illustrated in Figure~\ref{model-concepts.pdf}b: The input space is separated again from the output space, preventing models from fully leveraging knowledge acquired from solving other tasks.

Fine-tuning for text generation tasks keeps input and output in the same space; thereby, for example, connections between an argument and its improved version can be learned. Still, it faces a second restriction that is shown in Figure~\ref{model-concepts.pdf}b': The idea behind the mapping from input to output (why is an output correct for an input) remains fully implicit in the training pairs. For argument quality assessment, both restrictions imply that only those interdependencies between quality notions as well as those contextual influence factors are taken into account that are explicitly modeled or controlled by the human developer. Even though a lot of other interdependencies and factors may be well-known in argumentation theory \cite{vaneemeren:2004,aristotle:2007}, they are widely ignored thereby. This is where instruction-following LLMs go beyond.

\subsection{Instruction-Following LLMs} 

Instruction fine-tuning teaches LLMs to follow user instructions to solve tasks \cite{peng:2023}. As Figure~\ref{model-concepts.pdf}c stresses, it does so by representing the task in the joint space of inputs and outputs across all tasks and contexts, that is, in natural language (more precisely, in the embedding space in which language is encoded and from which it is decoded). Following the instruction fine-tuning paradigm, deriving an output from an input remains a language modeling problem as well as how to operationalize the derivation. This means that all knowledge ever processed in pretraining is still accessible in principle (bounded by the technical constraints of the model). Then, the supreme capability of transformers to integrate and leverage knowledge across contexts enables instruction-following LLMs to tackle unseen tasks with no or little fine-tuning. 

In argument quality assessment, we can expect that most knowledge about quality notions and their interdependencies as well as about influence factors and their effect on the subjective perception of argument quality has already been processed by leading LLMs, such as GPT-4 \cite{openai:2023} and Alpaca \cite{taori:2023}, in their pretraining stage. It should thus be possible to learn through instructions what is important in the assessment task at hand while not ignoring interactions with the surrounding concepts of argument quality.

\subsection{Instructions for Assessment}

Fine-tuning LLMs on general-purpose instruction data \cite{wang:2023a} will help them solve language modeling tasks in principle. By default, however, LLMs do not necessarily have access to what is to be prioritized for the setting of the task at hand. We expect that this is the information an LLM needs to be instructed with to assess argument quality reliably. Accordingly, we see the survey results from Section~\ref{sec:relatedwork} as an adequate basis to explore what to teach LLMs for the assessment. Instructions may thus include but are not limited to: 

\begin{itemize}
\setlength{\itemsep}{0pt}
\item
arguing goals, from agreement \cite{elbaff:2018} to deliberation \cite{falk:2021};
\item
definition of various quality notions, be it for maximal quality \cite{lauscher:2020} or minimal quality \cite{jin:2022};
\item
specificities of audiences \cite{lukin:2017} and debaters \cite{durmus:2019};
\item
background on controversial topics, such as other arguments \cite{luu:2019} or relationships between the topics \cite{zhao:2021};
\item
ethical aspects, such as biases \cite{holtermann:2022} and culture \cite{chen:2022};
\item
any examples of respective assessments, following the common few-shot learning idea.
\end{itemize}

Exemplarily, let us get back to the Huckleberry Finn claim from Section~\ref{sec:introduction} taken from kialo.com. On this online platform, users create and refine claims, and they give impact votes from 1 to 5. These impact votes may serve as gold labels. For voting, users are asked to consider both a claim's persuasiveness and its relevance to the claim it replies to, in equal weight.%
\footnote{https://support.kialo-edu.com/en/hc/about-voting/}
When using such labels, supervised learning (Figure~\ref{model-concepts.pdf}a) and standard transformer fine-tuning (Figure~\ref{model-concepts.pdf}b/b') can learn the semantics of the task only through the mapping from claim to vote, risking spurious correlations \cite{thorn-jakobsen:2021} as well as bias \cite{spliethoever:2020}. The equal weighting will likely not to be captured either,  as users may not weigh systematically, may not have read instructions, or may just have a subjective perception. 

Moreover, the same claim should certainly be assessed differently depending on the setting (beyond kialo.com). From a deliberative perspective, for example, it lacks a concrete counterproposal (e.g., \emph{even if we do not censor, a preface should be added or teachers should discuss the load of the N-word in everyday life}). Moving to quality improvement, the claim may need to be revised for specific audiences, for example, the concept of ``everyday midwest American English'' may be completely opaque to people with low literacy. 

Instructions may overcome all these challenges. An example may be:  \emph{``Rate the claim's quality from the perspective of deliberation, when presented to a person of low literacy''}. This makes the semantics of the task much more explicit, likely reducing biases and spurious correlations. It performs a stage-setting and an addressee-setting function which are crucial for assessment and improvement. We do not aim to come up with the best instructions in this paper, but see this as a task for future work. Rather, we discuss how~to~systematize respective instruction fine-tuning attempts.

\subsection{Blueprint for Instruction Fine-Tuning}

Effective processes for the general instruction fine-tuning of LLMs have already been established in prior work \cite{taori:2023}. Given the discussed advantages of such LLMs over previous models, we argue that argument quality assessment may be brought to the next level through systematic approaches to \emph{task-related} instruction fine-tuning. The idea is to bring specific knowledge about theories, circumstances, and ethical constraints of arguing along with ways of how to solve argument-related problems into the fine-tuning process. As a blueprint, such an approach could roughly consist of the following stages:
\begin{enumerate}
\setlength{\itemsep}{0pt}
\item
Start from a general instruction-following LLM, such as Alpaca \cite{taori:2023}. Even some standard pretrained transformer may suffice, if general instructions are added to Step~2.
\item
Acquire a seed set of argumentation-specific instructions, covering concepts such as those in the previous subsection. For example, these instructions can be derived manually or semi-automatically from the various datasets and experiments covered in the surveyed papers.
\item
Depending on available resources, apply techniques such as reinforcement learning using human feedback \cite{ouyang:2022}, fine-tuning on self-generated instructions \cite{wang:2023a}, or other instruction fine-tuning mechanisms that are proposed in research.
\item
Align the behavior of the instruction fine-tuned LLM on new unseen tasks at hand using systematic prompt design, for example, via soft prompting \cite{qin:2021} or sociodemographic prompting to emulate social profiles of debaters and audiences \cite{beck:2023}. Due to Step~3, these tasks now benefit from argument-specific task-solving skills.
\end{enumerate}

At least for fact-related argument quality dimensions, such \emph{local acceptability} \cite{wachsmuth:2017b}, an additional step may be to systematically work against hallucinations, by teaching the LLM to check arguments against some fact source (e.g., a knowledge base or a corpus). Factuality measures may be included in the model optimation for this purpose. We note, though, that this presupposes a setting in which sources can also be accessed at inference time. Besides, many quality dimensions are actually not (inherently) about facts, such as those from rhetorics \cite{wachsmuth:2017b}.

We expect that the resulting LLM will assess argument quality more reliably in line with the theories behind diverse quality notions and will adjust to the subjective viewpoints of interest. Thereby, various new opportunities emerge for real-world applications, as discussed in the next section.

\subsection{Evaluation of Quality Assessment}

Finally, we make a note on the evaluation of LLM-based argument quality assessment, to give a basic guideline. Various ways of evaluating assessment have been pursued in prior work; particularly, there is a debate about whether quality should be assessed in absolute terms, based on a given score range, or in relative terms, comparing different arguments to one another \cite{wachsmuth:2017d}. Instruction-following LLMs might not entirely resolve the issue behind; while they provide new means for a reliable evaluation (e.g., handling of context), their generative nature may also complicate the validation against some ground-truth. 

Ultimately, a fully \emph{unified} evaluation procedure may not be possible, because it depends on what information is available in a given assessment setting. Rather, we propose that an evaluation procedure \emph{ideally} takes into account the main decisive factors of quality, as we exemplified for the Huckleberry Finn claim above: What quality dimension is of interest, who is the audience of the argument, and similar. The evaluation may happen on a careful selection of existing datasets, but new benchmarks that account for the factors may be needed, too.

Criteria-wise, we see a mix of absolute and relative assessment as best approximating how humans assess quality, but this requires careful operationalization: Many argument quality dimensions imply some hard constraints, which speaks for an absolute part (e.g., are the argument's premise acceptable?). However, there may not be a clear best/worst quality for an argument, which speaks for a relative part wherever other arguments are accessible (e.g., are the premises more acceptable than those of other arguments?). Instruction fine-tuning should prepare an LLM for dealing with both parts and, hence, be evaluated against them.

 % About 1.50 pages
\section{Opportunities for the Real World}
\label{sec:analysis}

Arguably, instruction-following LLMs generally provide great opportunities for NLP and its applications. Their wide and easy applicability, along with their often low need for task-specific training data, is particularly beneficial in the context of interdisciplinary research. With a successful realization of the blueprint delineated above, however, we explicitly see specific potential for computational argumentation applications, due to their inherent need for argument quality assessment (see Section~\ref{sec:introduction}). We now sketch some of the main opportunities we see. Partly, they bring up ethical concerns, though, that we discuss at the end of the paper (Section~\ref{sec:limitations}).\,\,

\paragraph{Debating Technologies}

So far, one of the most impressive applications is IBM's Project Debater, which has competed well with professional human debaters \cite{slonim:2021}. Its quality assessment methods are audience-agnostic~\cite{toledo:2019,gretz:2020}, which may not suffice to convince people across diverse backgrounds, as research indicates \cite{alshomary:2022}. Moreover, Project Debater's arguments are retrieved, recomposed, and rephrased rather than written naturally. If controlled well, the generation capabilities of LLMs may advance notably on the latter, whereas our proposed fine-tuning process may explicitly target the adjustment to audiences. 

\paragraph{Argument Search}

Argument search aims to find the best pros and cons on controversial topics. Unlike for debating technologies, its goal to aid self-determined opinion formation suggests not to tune towards audiences. However, argument search engines miss a reliable quality-based ranking so far \cite{wachsmuth:2017e,stab:2018,dumani:2020}, likely due to the heterogeneity of argumentative domains and genres on the web. The low training need of instruction-following LLMs may alleviate this shortcoming. In addition, the text rewriting capabilities of LLMs may be employed to optimize the presentation of arguments \cite{skitalinskaya:2023b}, or to fill gaps as needed. We expect that convincing rankings and presentations are key to making people open to argument search, enabling them to overcome filter bubbles. 

\paragraph{Discussion Moderation}

The moderation of (online) content is critical to ensure healthy and productive discussions \cite{park:2012,park:2021,vecchi:2021}. This holds particularly for deliberative contexts, where participants should be supported in communicating their viewpoints.
%Any gain attained by LLMs in the analysis of argument quality based on relevance, coherence, logical structure, evidence usage, and overall persuasiveness will in turn provide invaluable insights for moderators in assessing the strength and effectiveness of arguments within online discussions.
Effective moderation reaches a bottleneck as the scale of online discussions grows \cite{klein:2012,shortall:2022}. LLMs instructed for argument quality can assist moderation efforts by detecting possible violations of community guidelines, inappropriate language, or generally low-quality arguments in discussions. This way, moderators can focus their attention on nuanced cases and appeals, optimizing efficiency and ensuring a healthier discourse. In some settings, generative LLMs could even lead a dialogue with users to provide clarifications, feedback, and improvement suggestions. 
%ultimately enhancing the quality of arguments in deliberative discourse. %By leveraging these capabilities of LLMs, we can advance the assessment of argument quality in online discussions, ultimately promoting healthier and more productive deliberative discourse.

\paragraph{Argumentative Writing Support} 

LLMs may further provide individualized education to learners (e.g., students or non-native speakers) as well as to everyday writers (e.g., e-commerce customers), for instance, by giving feedback on the quality of their arguments \cite{carlile:2018,chen:2022}. Instruction fine-tuning makes it easier to go beyond simple quality scoring (e.g., how clear an argument is) to targeted hints (e.g., \emph{Provide more evidence for your initial claim!}). Prompted with the writing goal, LLMs may also suggest argument completions, such as missing conclusions \cite{gurcke:2021}. With these means, students may learn to reason more soundly, product reviews can become more informative, and so forth. LLMs instructed with the concrete feedback scenario (e.g., \emph{a student learning to write essays in English}) will help to further individualize support and may even adjust to the specific learning need of the user.

\paragraph{Other Applications} 

% NOTE (HW). Left out for lack of literature support
%One example is \textit{reading support}: The idea is to recommend part of the text (i.e., recommend argument) based on their qualities which are particularly relevant for a certain user. These qualities may be linked to the form (i.e., the genre, or the used language level) or to the content (i.e., the topic, a precise viewpoint which is conveyed). These tools can be used also to address argument-based summarization of text or discussions, through the selection in the summary of ``high" quality arguments.

In several other scenarios, argument quality is crucial. One example is to \emph{generate summaries} of the best arguments in news articles \cite{syed:2020} or online discussions \cite{syed:2023}. Here, instruction-following LLMs can interpret the term \emph{best} as needed---without any task-specific fine-tuning. In the \emph{medical domain}, argument quality plays a central role for evidence-based medicine \cite{mayer:2021}. A well-instructed LLM may assess evidence strength, thus enabling better inferences based on clinical trials or reports. Similarly, the reasonableness of arguments on health discussion online platforms may be evaluated. 
% NOTE (HW). Left out for lack of further literature support
%In this case, the assessment of the argument quality do not concern only the assessment of the quality of the argument components but, in particular, the assessment of the reasonableness of the addressed argumentation reasoning. In the medical scenario, it is also important to generate high quality explanations behind a certain diagnosis to train medical residents to justify the selected diagnosis and treatment. These high-quality medical explanations need to rely on well assessed arguments. 
Further scenarios include \emph{e-commerce}. There, an LLM-based service chatbot can, for example, select arguments based on quality notions (e.g., clarity) to explain to customers why a request cannot be completed, to minimize their dissatisfaction. Argument quality may also~be assessed in \emph{recommender systems} to make justified suggestions based on compelling reasons. 

\paragraph{Implications for Research} 

Finally, we also see great potential for diversity and subjectivity-aware instruction fine-tuning when it comes to driving fundamental research, as sketched here for two examples: interdisciplinary work at the interface of NLP and computational social science, and the methodological development driven by the need to cope with subjectivity in argument quality annotations.

The social science context adds even more \emph{diversity}, including sophisticated quality notions and domain-specific language, along with new challenges, such as well-curated and annotated, but small and imbalanced datasets \cite{falk:2022}. Our instruction fine-tuning blueprint fits exactly such scenarios: annotation guidelines serve as instructions, highly-curated annotations as reinforcement examples, and the knowledge encoded in LLMs alleviates resource-lean issues. %Moreover encoding the output (the annotation schema) in a richer space shared with  the input can support the models to account for the hierarchical coding schemes and unbalanced annotations discussed above. 
Additionally, the scene-setting function of instruction fine-tuning (see Section~\ref{sec:model}) has the potential to address the deliberative goal of defining and quantifying discourse quality across contexts \cite{esau:2021}. 

The multiple factors of \textit{subjectivity} influencing argument quality perception (debater and audience beliefs, values, etc.) often limit the inter-annotator agreement \cite{wachsmuth:2017b}. Ultimately, subjectivity is a constitutive feature of argument quality, as indicated above. \newcite{romberg:2022} suggest to join the perspectivist turn of machine learning and NLP \cite{plank:2022,cabitza:2023} in computational argumentation. LLMs' perspective-taking capabilities could be a game changer for this, assuming that the risks of sociodemographic prompting \cite{beck:2023} and stereotypes \cite{cheng:2023} are properly dealt with. 
 % About 1.00 pages
\section{Conclusion}
\label{sec:conclusion}

Argument quality assessment has become a core task in NLP research on computational argumentation, due to its importance for various applications, from debating technologies and argument search to discussion moderation and writing support. However, a reliable assessment is often hampered by the diversity of quality notions involved and the subjectivity of their perception. In this survey-based position paper, we have raised the question of how to drive research on instruction-following large language models (LLMs) for argument quality to substantially evolve the state of the art.

Our survey of 83 recent papers confirms that argument quality research often targets conceptual quality notions and the factors that influence these notions, aside from the computational assessment and improvement of argument quality. We have argued that many limitations of prior work can be overcome, if LLMs are not just simply prompted for argument quality assessment, but if systematic ways to instruct LLMs for argument quality during instruction fine-tuning are found. This is due to the fact that instruction-following LLMs, for the first time in machine learning-based NLP research, make the connection between inputs and outputs of tasks explicit, namely, through the instructions. Thereby, all knowledge that an LLM has processed during pretraining and fine-tuning can be shared across tasks and contexts.

To guide future work in this direction, we have delineated a blueprint of how to approach the instruction fine-tuning process. Realizations of this process will likely bring up further~problems, not all are foreseeable at this point. Moreover, LLMs that effectively predict human perception of argument quality directly raise concerns, as detailed in our ethics statement below. Still, we are confident that coordinated efforts towards sustainable research on LLMs for argument quality will enable the community to progress on core visions of computational argumentation---whether it is about ways to overcome filter bubbles or about the individualized support of argumentation learners. The paper at hand seeks to lay the ground for this research.\,\,\,

  % About 0.25 pages

\section{Ethics Statement}
\label{sec:limitations}

Despite the huge potential of instruction-following LLMs for argument quality assessment across various applications of computational argumentation outlined in Section~\ref{sec:analysis}, the blueprint from Section~\ref{sec:model} also comes with limitations and ethical concerns. We acknowledge and analyze these in this section.\,\,
% NOTE. Doesn't really work with the page restrictions
%%%\footnote{Due to the inherence of the issues discussed here in the topic of this position paper, we plan to integrate this section in the main body of the paper upon acceptance.}

\subsection{Limitations}

The discussed potential we see is based on our survey of argument quality research (Section~\ref{sec:relatedwork}), initial works of the emerging body of instruction fine-tuning research~\citep[e.g.,][]{peng:2023}, and our own preliminary tests. Yet, the work at hand remains a position paper, meaning that experimental research still needs to establish whether the outlined blueprint or similar paths will actually result in substantial progress.  It is possible that argument-specific instruction fine-tuning of large language models (LLMs) does not improve over the capabilities of a general large-scale tuning. Also, the systematic ways that we have proposed to establish above remain to be found; there is no obvious way of directly obtaining them. This challenge is in line with the overall state of instruction fine-tuning research, both in academia and in industry.\,\,\,

Regarding the specific challenges of argument quality raised in Section~\ref{sec:introduction}, another limitation refers to general possibility that information required to achieve a realiable assessment is simply not available, due to specificities of the setting or underlying privacy regulations. This particularly includes the audience whose quality perception is to be represented, but possibly also aspects of the (temporal, geographical, and social) context in which an argument is to be considered. Also, as soon as we rely on human-created training data for instruction fine-tuning, the creators' biases and values affect its impact. Ultimately, we cannot expect LLMs to tackle a task reliably under conditions that simply do not suffice to make an informed judgment.

\subsection{Ethical Concerns}

Many arising ethical issues of the use of LLMs for argument quality assessment are general and not specific to computational argumentation, such as the increased environmental impact of bigger models, privacy issues, hallucinations, the potential of models to encode unfair exclusive~\cite{dev:2021,lauscher:2022} and stereotypical biases~\cite{blodgett:2020}, which may result in allocational and representational harms~\citep{barocas:2017}. However, we believe that some of them deserve specific attention in scenarios where argument quality is assessed or optimized, particularly when leveraging the power of LLMs.

In particular, argument quality assessment may be used in sensitive applications such as digital education; for example, to support argumentative writing or to provide guidance on political opinion formation. For such applications, factual errors are particularly problematic, as they may easily lead to wrong or shifted beliefs. Whenever LLMs may generate argumentative content, say, for debating technologies or to fill gaps in argument search as discussed, extra measures should thus be taken to prevent hallucinations. We have sketched how to generally account for them in Section~\ref{sec:model}, but fully avoiding them may be hard given how LLMs work.\,\,\,

Similarly, unfair social biases are easy to perpetuate in such applications, since the output of LLMs for argument quality assessment will often directly affect human views. This raises various integral and partly self-referential questions, such as who decides on what makes a good argument, or, how to decide on the ethical uses of instruction-following LLMs in argument quality assessment? We expect that universally-accepted answers to these questions may not exist, as they also depend on the values within a culture or society.

As the limitations discussed above imply, further ethical concerns refer to the tension between the inclusion of audience and debater information for a more accurate  quality assessment. While an argument's persuasive effect is, for instance, highly dependent on the sociodemographic aspects of its audience, it is questionable in general to what extent an application of respective methods should have access to personal data. Such aspects need to be handled with care, and under consultation with an ethics board, where needed.

For a successful and societal beneficial use of instruction-following LLMs, we thus conclude that future research on argument quality assessment needs to find answers to such questions and to proactively raise and discuss them explicitly.

  % About 0.75 pages
\section{Acknowledgments}

This work was partially funded by the Deutsche Forschungsgemeinschaft (DFG) within the project OASiS, project number 455913891, as part of the Priority Program ``Robust Argumentation Machines (RATIO)'' (SPP-1999), as well as within the project ArgSchool, project number 453073654. It was also partially funded by the Bundesministerium f\"ur Bildung und Forschung (BMBF) within the project E-DELIB, project number 01IS20050, and under the Excellence Strategy of the German Federal Government and the States.  
This work has also been supported by the French government, through the 3IA C\^ote d'Azur Investments in the Future project managed by the National Research Agency (ANR) with the reference number ANR- 19-P3IA-0002.  

\section{Bibliographical References}\label{sec:reference}
\bibliographystyle{lrec-coling2024-natbib}
\bibliography{coling24-argument-quality-lit}

\appendix
\section{List of Surveyed Papers}
\label{sec:survey}

Table~\ref{sec:survey} shows the full list of all 83 surveyed papers that resulted from our search and filtering process in Section~\ref{sec:relatedwork}, along with the primary research direction and main aspect of each paper and the research community where the respective paper has been published: natural language processsing (NLP), artificial intelligence (AI), information retrieval (IR), or computational argumentation (CA). The research directions and aspects are detailed in Section~\ref{sec:relatedwork} as well as selected papers from the list.

\begin{table*}[t]
\scriptsize
\renewcommand{\arraystretch}{0.925}
\setlength{\tabcolsep}{10pt}
\centering
\begin{tabular*}{0.95\linewidth}{rllll}
\toprule
% 		& \multicolumn{2}{c}{\bf  }	&& \multicolumn{2}{r}{\bf  }  \\
%		\cmidrule(l@{4pt}r@{0pt}){2-3}		\cmidrule(l@{4pt}r@{0pt}){5-6}	
%\bf  	& \bf         &    			&& \bf  	   & \bf  		 \\	
\bf \#	& \bf Research Direction	& \bf Main Aspect			& \bf Paper	& \bf Community	\\
\midrule		
1     	& \bf Conceptual Notions	& Notions of Maximal Quality	& \newcite{atkinson:2019}	&	NLP	\\
2	&					&						& \newcite{beigman-klebanov:2017}	&	NLP	\\
3	&					&						& \newcite{dumani:2020}	&	IR	\\           				
4     	& 	           			& 	           				&\newcite{elbaff:2018}	&	NLP	\\
5     	& 					& 	           				& \newcite{falk:2021}	&	NLP	\\
6     	& 					& 	           				& \newcite{gretz:2020}	&	AI	\\
7     	& 					& 						& \newcite{guo:2023} & AI \\
8     	& 	           			& 	           				& \newcite{joshi:2023}	&	NLP	\\
9     	& 					& 	           				& \newcite{ke:2019}	&	NLP	\\
10     & 					& 	           				&\newcite{kolhatkar:2017b}	&	NLP	\\
11    	& 					& 	           				&\newcite{lauscher:2020}	&	NLP	\\
12   	& 					& 						& \newcite{ng:2020}	&	NLP	\\
13   	& 	           			&            					& \newcite{nguyen:2018}	&	AI	\\
14   	&					& 	          				& \newcite{potthast:2019}	&	IR	\\
15   	& 					& 	           				& \newcite{shiota:2022}	&	NLP	\\
16   	& 	           			&            					&\newcite{wachsmuth:2017d}	&	NLP	\\
\addlinespace
17     & 	           			& Notions of Minimal Quality	& \newcite{habernal:2018}	&	NLP	\\
18     & 	           			& 	           				&\newcite{haworth:2021}	&	AI	\\
19     & 	           			& 	           				& \newcite{kolhatkar:2017}	&	NLP	\\
20     & 	           			& 	           				&\newcite{jin:2022}	&	NLP	\\
21     & 	           			& 	           				&\newcite{park:2018}	&	NLP	\\
22   	& 					& 	           				& \newcite{pauli:2022}	&	NLP	\\
23     & 	           			& 						&\newcite{persing:2017}	&	AI	\\
24	& 	           			& 	           				&\newcite{ziegenbein:2023}	&	NLP	\\
\midrule
25   	& \bf Influence Factors	& Argument-related Factors 	&\newcite{carlile:2018}	&	NLP	\\
26   	& 			           	& 	           				&\newcite{chen:2022}	&	NLP	\\
27   	& 	           			&             					& \newcite{durmus:2019b}	&	NLP	\\
28   	& 	           			&             					& \newcite{egawa:2019}	&	NLP	\\
29	& 	           			& 	           				&\newcite{elbaff:2020a}	&	NLP	\\
30     & 			 	         & 						&\newcite{kobbe:2020}	&	NLP	\\
31   	& 	           			&             					& \newcite{li:2020}	&	NLP	\\
32   	& 	      				& 	          		 		&\newcite{luu:2019}	&	NLP	\\
33   	& 	          			&             					& \newcite{persing:2017a}	&	NLP	\\
34   	& 	           			& 	           				& \newcite{potash:2017}	&	NLP	\\
35   	& 	           			&             					& \newcite{skitalinskaya:2021}	&	NLP	\\
36   	& 	           			&             					& \newcite{toledo:2019}	&	NLP	\\
37   	& 	           			& 	           				&\newcite{wachsmuth:2020}	&	NLP	\\
38   	& 	           			&             					&\newcite{wang:2017}	&	NLP	\\
39   	& 	           			&             					&\newcite{wambsganss:2022}	&	NLP	\\
40   	&           				&     						&\newcite{zhao:2021}	&	NLP	\\
\addlinespace
41     & 	           			& Context-related Factors     	& \newcite{alkhatib:2020}	&	NLP	\\
42     & 	           			&             					&\newcite{alshomary:2022}	&	NLP	\\
43     & 	           			&             					&\newcite{durmus:2018}	&	NLP	\\
44     & 	           			&             					& \newcite{durmus:2019}	&	NLP	\\
45     & 	           			&             					&\newcite{elbaff:2020b}	&	NLP	\\
46     & 	           			&             					&\newcite{fromm:2023}	&	NLP	\\
47     & 	           			&             					&\newcite{gu:2018}	&	NLP	\\
48     & 	           			&             					&\newcite{hasan:2021}	&	NLP	\\
49     &            				& 						& \newcite{kornilova:2022}	&	NLP	\\
50     &           				&           					&\newcite{jain:2021}	&	NLP	\\
51     &             				&             					&\newcite{liu:2022}	&	NLP	\\
52	& 	           			&            	 				& \newcite{longpre:2019}	&	NLP	\\
53     & 	           			&             					& \newcite{lukin:2017}	&	NLP	\\
54     & 	           			&             					& \newcite{wiegmann:2022}	&	NLP	\\
\midrule
55    	& \bf Computational Models & Models for Assessment 	& \newcite{ding:2023}	&	NLP	\\
56   	&             				&             					&\newcite{falk:2022}	&	NLP	\\
57   	&             				&             					& \newcite{falk:2023}	&	NLP	\\
58    	&             				&             					&\newcite{feger:2020}	&	CA	\\
59    	&             				&             					& \newcite{gienapp:2020}	&	NLP	\\
69   	&             				&             					& \newcite{gleize:2019}	&	NLP	\\
61   	&             				&             					& \newcite{gurcke:2021}	&	NLP	\\
62    	&             				&             					&\newcite{holtermann:2022}	&	NLP	\\
63	&             				&             					& \newcite{huang:2021}	&	AI	\\
64	&             				&             					& \newcite{jo:2018}	&	NLP	\\
65   	&             				&             					&\newcite{kondo:2021}	&	NLP	\\
66   	&             				&             					& \newcite{liu:2021}	&	NLP	\\
67    	&             				&             					& \newcite{marro:2022}	&	NLP	\\
68    	&             				&             					&\newcite{mim:2019}	&	NLP	\\
69   	&             				&             					&\newcite{saveleva:2021}	&	NLP	\\
70    	&             				&           					&\newcite{simpson:2018}	&	NLP	\\
71   	& 					& 						&\newcite{song:2020}	&	AI	\\
72   	&             				&             					& \newcite{vandermeer:2022}	&	NLP	\\
\addlinespace
73   	&             				& Models for Improvement  	&\newcite{ke:2018}	&	AI	\\
74   	&             				&             					&\newcite{skitalinskaya:2023}	&	NLP	\\
75   	&             				&            					&\newcite{skitalinskaya:2023b}	&	NLP	\\
\midrule
76   	&  \bf Other 			& Other					&\newcite{chalaguine:2020}	&	CA	\\
77	&             				&             					&\newcite{chen:2022b}	&	NLP	\\
78    	&             				&             					&\newcite{fromm:2021}	&	AI	\\
79   	&		            		& 						&\newcite{heinisch:2022}	&	NLP	\\
80   	&            				&            					&\newcite{kees:2021}	&	NLP	\\
81   	&             				&             					& \newcite{kiesel:2020}	&	IR	\\
82   	&             				&             					&\newcite{rach:2020}	&	NLP	\\
83    	&             				&             					&\newcite{yang:2019}	&	NLP	\\
\bottomrule
\end{tabular*}
\caption{The list of all 83 surveyed NLP, AI, IR, and CA papers, ordered by their primary general research direction and main aspect and then by author names and year. 23 papers deal with \emph{conceptual notions} primarily, 31 with \emph{influence factors}, 21 with \emph{computational models}, and eight have \emph{other} primary directions.} 
\label{table-survey}
\end{table*}

\end{document}